\documentclass{article} % For LaTeX2e
\usepackage{iclr2017_conference,times}
\usepackage{hyperref}
\usepackage{url}
\usepackage{amsmath}
\usepackage{amsthm}
\usepackage{amssymb}
\usepackage{amsfonts,eucal,amsbsy}
\usepackage{mathtools}
\usepackage{latexsym}
\usepackage{graphicx}
\usepackage{epstopdf}
\usepackage{longtable}
\usepackage{tabularx}
\usepackage{amsfonts}
\usepackage{enumerate}
\usepackage{tikz}
\usepackage{float}
\usepackage{fancyvrb}
\usepackage{pdflscape}
\usepackage{multirow}
\usepackage{enumitem}
\usepackage{booktabs}
\usepackage{algorithm}
\usepackage{algpseudocode}
\usepackage{xr}
\usepackage{subcaption}
\usepackage{CJKutf8}

\DeclareMathOperator*{\argmax}{arg\,max}

\DeclareMathOperator*{\topk}{topk}
\DeclareMathOperator*{\argtopk}{arg\,topk}
\DeclareMathOperator*{\candidate}{getCandidateOutputs}

\title{The Neural Noisy Channel}

\author{Lei Yu$^1$\thanks{Work completed at DeepMind.} , Phil Blunsom$^{1,2}$, Chris Dyer$^{2}$, Edward Grefenstette$^{2}$, and Tom\'{a}\v{s} Ko\v{c}isk\'{y}$^{1,2}$ \\
$^1$University of Oxford and $^2$DeepMind \\
{\tt lei.yu@cs.ox.ac.uk, \{pblunsom,cdyer,etg,tkocisky\}@google.com}
}

\iclrfinalcopy % Uncomment for camera-ready version

\begin{document}

\maketitle

\begin{abstract}
We formulate sequence to sequence transduction as a noisy channel decoding problem and use recurrent neural networks to parameterise the source and channel models. Unlike direct models which can suffer from explaining-away effects during training, noisy channel models must produce outputs that explain their inputs, and their component models can be trained with not only paired training samples but also unpaired samples from the marginal output distribution. Using a latent variable to control how much of the conditioning sequence the channel model needs to read in order to generate a subsequent symbol, we obtain a tractable and effective beam search decoder. Experimental results on abstractive sentence summarisation, morphological inflection, and machine translation show that noisy channel models  outperform direct models, and that they significantly benefit from increased amounts of unpaired output data that direct models cannot easily use.
\end{abstract}

\section{Introduction}
Recurrent neural network sequence to sequence models~\citep{kalchbrenner:2013,sutskever:2014,bahdanau:2015} are excellent models of $p(\textrm{output sequence }\boldsymbol{y} \mid \textrm{input sequence }\boldsymbol{x})$, provided sufficient input--output $(\boldsymbol{x},\boldsymbol{y})$ pairs are available for estimating their parameters. However, in many domains, vastly more unpaired output examples are available than input--output pairs (e.g., transcribed speech is relatively rare although non-spoken texts are abundant; Swahili--English translations are rare although English texts are abundant; etc.). A classic strategy for exploiting both kinds of data is to use Bayes' rule to rewrite $p(\boldsymbol{y} \mid \boldsymbol{x})$ as $p(\boldsymbol{x} \mid \boldsymbol{y}) p(\boldsymbol{y})/p(\boldsymbol{x})$, a factorisation which is called a \textbf{noisy channel model}~\citep{shannon:1948}. A noisy channel model thus consists of two component models: the conditional \textbf{channel model}, $p(\boldsymbol{x} \mid \boldsymbol{y})$, which characterizes the \emph{reverse} transduction problem and whose parameters are estimated from the paired $(\boldsymbol{x},\boldsymbol{y})$ samples, and the unconditional \textbf{source model}, $p(\boldsymbol{y})$, whose parameters are estimated from both the paired and (usually much more numerous) unpaired samples.\footnote{We do not model $p(\boldsymbol{x})$ since, in general, we will be interested in finding $\argmax_\mathbf{y}p(\mathbf{y} \mid \mathbf{x})$, and $\argmax_\mathbf{y}p(\mathbf{y} \mid \mathbf{x}) = \argmax_\mathbf{y} \frac{p(\mathbf{x} \mid \mathbf{y})p(\mathbf{y})}{p(\mathbf{x})}= \argmax_\mathbf{y} p(\mathbf{x} \mid \mathbf{y})p(\mathbf{y})$.}

Beyond their data omnivorousness, noisy channel models have other benefits. First, the two component models mean that two different aspects of the transduction problem can be addressed independently. For example, in many applications, source models are language models and innovations in these can be leveraged to obtain improvements in any system that uses them as a component. Second, the component models can have complementary strengths, since inference is carried out in the product space; this simplifies design because a single model does not have to get everything perfectly right. Third, the noisy channel operates by selecting outputs that both are \emph{a priori} likely \emph{and} that explain the input well. This addresses a failure mode that can occur in conditional models in which inputs are ``explained away'' by highly predictive output prefixes, resulting in poor training \citep{klein:2001}. Since the noisy channel formulation requires its outputs to explain the observed input, this problem is avoided.

In principle, the noisy channel decomposition is straightforward; however, in practice, decoding (i.e., computing $\arg \max_{\boldsymbol{y}} p(\boldsymbol{x} \mid \boldsymbol{y}) p(\boldsymbol{y})$) is a significant computational challenge, and tractability concerns impose restrictions on the form the component models can take. To illustrate, an appealing parameterization would be to use an attentional seq2seq network \citep{bahdanau:2015} to model the channel probability $p(\boldsymbol{x} \mid \boldsymbol{y})$. However, seq2seq models are designed under the assumption that the complete conditioning sequence is available before any prefix probabilities of the output sequence can be computed. This assumption is problematic for channel models since it means that a complete output sequence must be constructed before the channel model can be evaluated (since the channel model conditions on the output). Therefore, to be practical, the channel probability must decompose in terms of prefixes of the conditioning variable, $\boldsymbol{y}$. While the chain rule justifies decomposing output variable probabilities in terms of successive extensions of a partial prefix, no such convenience exists for conditioning variables, and approximations must be introduced.

In this work, we use a variant of the newly proposed online seq2seq model of \citet{yu:2016} which uses a latent alignment variable to enable its probabilities to factorize in terms of prefixes of both the input and output, making it an appropriate channel model~(\S\ref{sec:model}). Using this channel model, the decoding problem then becomes similar to the problem faced when decoding with direct models~(\S\ref{sec:decoding}). Experiments on abstractive summarization, machine translation, and morphological inflection show that the noisy channel can significantly improve performance and exploit unpaired output training samples and that models that combine the direct model and a noisy channel model offer further improvements still~(\S\ref{sec:experiments}).

\section{Background: Segment to Segment Neural Transduction}
\label{sec:model}

Our model is based on the Segment to Segment Neural Transduction model (SSNT) of Yu et al., 2016. At a high level, the model alternates between encoding more of the input sequence and decoding output tokens from the encoded representation. This presentation  deviates from the Yu et al.'s presentation so as to emphasize the incremental construction of the conditioning context that is enabled by the latent variable.

\subsection{Model description}
Similar to other neural sequence to sequence  models, SSNT models the conditional probability $p(\boldsymbol{y} \mid \boldsymbol{x})$ of a output sequence $\boldsymbol{y}$ given a input sequence $\boldsymbol{x}$.

To avoid having to observe the complete input sequence $\boldsymbol{x}$ before making a prediction of the beginning of the output sequence, we introduce a latent alignment variable $\boldsymbol{z}$ which indicates when each token of the output sequence is to be generated as the input sequence is being read. Since we assume that the input is read just once from left to right, we restrict $\boldsymbol{z}$ to be a monotonically increasing alignment (i.e., $z_{j+1} \ge z_j$ is true with probability 1),
where $z_j = i$ denotes that the output token at position $j$ ($y_j$) is generated when the input sequence up through position $i$ has been read. The SSNT model is:
\begin{align}
\begin{split}
 p(\boldsymbol{y} \mid \boldsymbol{x}) & = \sum_{\boldsymbol{z}} p(\boldsymbol{y}, \boldsymbol{z} \mid \boldsymbol{x}) \\
 p(\boldsymbol{y}, \boldsymbol{z} \mid \boldsymbol{x}) & \approx   \prod_{j=1}^{|\boldsymbol{y}|} \underbrace{p(z_j \mid z_{j-1},
 \boldsymbol{x}_{1}^{z_j},
 \boldsymbol{y}_1^{j-1})}_{\text{alignment probability}} \underbrace{p(y_j \mid \boldsymbol{x}_{1}^{z_j},
 \boldsymbol{y}_1^{j-1})}_{\text{word probability}}. \label{eq:model}
\end{split}
\end{align}

We explain the model in terms of its two components, starting with the word generation term. In the SSNT, the input and output sequences $\boldsymbol{x}$, $\boldsymbol{y}$ are encoded with two separate LSTMs \citep{hochreiter1997long}, resulting in sequences of hidden states representing prefixes of these sequences. In Yu et al.'s  formulation, the input sequence encoder (i.e., the conditioning context encoder) can either be a unidirectional or bidirectional LSTM, but here we assume that it is a unidirectional LSTM, which ensures that it will function well as a channel model that can compute probabilities with incomplete conditioning contexts (this is necessary since, at decoding time, we will be constructing the conditioning context incrementally). Let $\mathbf{h}_i$ represent the input sequence encoding for the prefix $\boldsymbol{x}_1^{i}$. Since the final action at timestep $j$ will be to predict $y_j$, it is convenient to let $\mathbf{s}_j$ denote the hidden state that excludes $y_j$, i.e., the encoding of the prefix $\boldsymbol{y}_1^{j-1}$.

The probability of the next token $y_j$ is calculated by concatenating the aligned hidden state vectors $\mathbf{s}_j$ and $\mathbf{h}_{z_j}$ followed by a softmax layer,
\begin{align*}
p(y_j \mid \boldsymbol{x}_1^{z_j}, \boldsymbol{y}_1^{j-1}) \propto \text{exp} (\mathbf{W}_w[\mathbf{h}_{z_j};\mathbf{s}_j] + \mathbf{b}_w).
\end{align*}
The model thus depends on the current alignment position $z_j$, which determines how far into $\boldsymbol{x}$ it has read.

We now discuss how the sequence of $z_j$'s are generated. First, we remark that modelling this distribution requires some care so as to avoid conditioning on the entire input sequence. To illustrate why one might induce a dependency on the entire input sequence in this model, it is useful to compare to a standard attention model. Attention models operate by computing a score using a representation of alignment candidate (in our case, the candidates would be every unread token remaining in the input). If we followed this strategy, it would be necessary to observe the full input sequence when making the first alignment decision.

We instead model the alignment transition from timestep $j$ to $j+1$ by decomposing it into a sequence of conditionally independent \textsc{shift} and \textsc{emit} operations that progressively decide whether to read another token or stop reading. That is, at input position $i$, the model decides to \textsc{emit}, i.e., to set $z_j=i$ and predict the next output token $y_j$ from the word model, or it decides to \textsc{shift}, i.e., to read one more input token and increment the input position $i \gets i+1$. The probability $p(a_{i,j} = \textsc{emit} \mid \boldsymbol{x}_1^{i}, \boldsymbol{y}_1^{j-1})$ is calculated using the encoder and decoder states defined above as:
\begin{align*}
p(a_{i,j} = \textsc{emit} \mid \boldsymbol{x}_{1}^{i}, \boldsymbol{y}_1^{j-1}) = \sigma(\text{MLP}(\mathbf{W}_t[\mathbf{h}_{i};\mathbf{s}_j] + b_t)).
\end{align*}
The probability of \textsc{shift} is simply $1-p(a_{i,j} = \textsc{emit})$. In this formulation, the probabilities of aligning $z_j$ to each alignment candidate $i$ can be computed by reading just $\boldsymbol{x}_1^i$ (rather than the entire sequence). The probabilities are also independent of the contents of the suffix $\boldsymbol{x}_{i+1}^{|\boldsymbol{x}|}$.

Using the probabilities of the auxiliary $a_{i,j}$ variables, the alignment probabilities needed in Eq.~\ref{eq:model} are computed as:
\begin{align*}
p(z_j = i \mid z_{j-1}, \boldsymbol{y}_1^{j-1}, \boldsymbol{x}_{1}^{i}) &= \begin{cases}
0 & \text{if }i < z_{j-1} \\
p(a_{i,j} = \textsc{emit}) & \text{if }i=z_{j-1} \\
\left(\prod_{i'=z_{j-1}}^{i-1} p(a_{i',j} = \textsc{shift}) \right) p(a_{i,j} = \textsc{emit}) & \text{if }i>z_{j-1}
\end{cases}
\end{align*}

\subsection{Inference algorithms}
In SSNT, the probability of generating each $y_j$ depends only on the current output position's alignment ($z_j$), the current output prefix ($\boldsymbol{y}_1^{j-1}$), the input prefix up to the current alignment ($\boldsymbol{x}_1^{z_j}$). It does \emph{not} depend on the history of the alignment decisions. Likewise, the alignment decisions at each position are also conditionally independent of the history of alignment decisions. Because of these  independence assumptions, $\boldsymbol{z}$ can be marginalised using a $O(|\boldsymbol{x}|^2 \cdot |\boldsymbol{y}|)$ time dynamic programming algorithm where each fill in a chart with computing the following marginal probabilities:
\begin{align*}
\begin{split}
 \alpha(i,j) & = p(z_j=i, \boldsymbol{y}_1^j \mid \boldsymbol{x}_1^{z_j}) = \sum_{i'=1}^{i} \alpha(i',j-1)  \underbrace{p(z_j \mid z_{j-1},
 \boldsymbol{x}_{1}^{z_j},
 \boldsymbol{y}_1^{j-1})}_{\text{alignment probability}} \underbrace{p(y_j \mid \boldsymbol{x}_{1}^{z_j},
 \boldsymbol{y}_1^{j-1})}_{\text{word probability}}.
\end{split}
\end{align*} 

The model is trained to minimize the negative log likelihood of the parallel corpus $S$:
\begin{equation}
\label{loss}
\begin{split}
\mathcal{L}(\boldsymbol{\theta}) &= - \sum_{(\boldsymbol{x}, \boldsymbol{y}) \in S} \log p(\boldsymbol{y}\ |\ \boldsymbol{x}; \boldsymbol{\theta})\\
&= - \sum_{(\boldsymbol{x}, \boldsymbol{y}) \in S} \log \alpha(|\boldsymbol{x}|, |\boldsymbol{y}|).\\ 
\end{split}
\end{equation}
The gradients of this objective with respect to the component probability models can be computed using automatic differentiation or using a secondary dynamic program that computes `backward' probabilities. We refer the reader to Section 3.1 of \citet{yu:2016} for details.

In this paper, we use a slightly different objective from the one described in \citet{yu:2016}. Rather than marginalizing over the paths that end in any possible input positions $\sum_{i=1}^I\alpha(i, |\boldsymbol{y}|)$, we require that the full input be consumed when the final output symbol is generated. This constraint biases away from predicting outputs without explaining them using the input sequence.

\section{Decoding}
\label{sec:decoding}
We now turn to the problem of decoding, that is, of computing
\begin{align*}
    \hat{\boldsymbol{y}} = \arg \max_{\boldsymbol{y}} p(\boldsymbol{x} \mid \boldsymbol{y}) p(\boldsymbol{y}),
\end{align*}
where we are using the SSNT model described in the previous section as the channel model and a language model that delivers prior probabilities of the output sequence in left-to-right order, i.e., $p(y_i \mid \boldsymbol{y}^{i-1})$.

Marginalizing the latent variable during search is computationally hard \citep{simaan:1996}, and so we approximate the search problem as
\begin{align*}
    \hat{\boldsymbol{y}} = \arg \max_{\boldsymbol{y}} \max_{\boldsymbol{z}} p(\boldsymbol{x},\boldsymbol{z} \mid \boldsymbol{y}) p(\boldsymbol{y}).
\end{align*}
However, even with this simplification, the search problem remains nontrivial. On one hand, we must search over the space of all possible outputs with a model that makes no Markovian assumptions. This is similar to the decoding problem faced in standard seq2seq transducers.  On the other hand, our model computes the probability of the given input conditional on the predicted output hypothesis. Therefore, instead of just relying on a single softmax to provide a probability for every output word type (as we conveniently can in the direct model), we must loop over each output word type, and run a softmax over the input vocabulary---a computational expense that is quadratic in the size of the vocabulary!

To reduce this computational effort, we make use of an auxiliary direct model $q(\boldsymbol{y}, \boldsymbol{z} \mid \boldsymbol{x})$ to explore probable extensions of partial hypotheses, rather than trying to perform an exhaustive search over the vocabulary each time we extend an item on the beam.

Algorithm~\ref{decode}, in Appendix~\ref{sec:algo_appendix}, describes the decoding algorithm based on a formulation by \citet{tillmann1997dp}. The idea is to create a matrix $Q$ of partial hypotheses. Each hypothesis in cell $(i,j)$ covers the first $i$ words of the input ($\boldsymbol{x}_1^i$) and corresponds to an output hypothesis prefix of length $j$ ($\boldsymbol{y}_1^j$). The hypothesis is associated with a model score. For each cell $(i, j)$, the direct proposal model first calculates the scores of possible extensions of previous cells that could then reach $(i,j)$ by considering every token in the output vocabulary, from all previous candidate cells $(i-1,\le j)$. That gives the top $K_1$ partial output sequences. These partial output sequences are subsequently rescored by the noisy channel model, and the $K_2$ best candidates are kept in the beam and used for further extension. The beam size $K_1$ and $K_2$ are hyperparameters to be tuned in the experiments. 

\subsection{Model combination}
The decoder we have just described makes use of an auxiliary decoding model.
This means that, as a generalisation, it is capable of decoding under an objective that is a linear combination of the direct model, channel model, language model and a bias for the output length\footnote{In the experiments, we did not marginalize the probability of the direct model when calculating the general search objective. We found that marginalizing the probability does not give better performance and makes decoding extremely slow.},
\begin{equation}
O_{\boldsymbol{x}_1^i, \boldsymbol{y}_1^j} = \lambda_1 \log p(\boldsymbol{y}_1^j\ |\ \boldsymbol{x}_1^i) + \lambda_2 \log p(\boldsymbol{x}_1^i\ |\ \boldsymbol{y}_1^j) + \lambda_3 \log p(\boldsymbol{y}_1^j) + \lambda_4 |\boldsymbol{y}_1^j|.
\end{equation}
The bias is used to penalize the noisy channel model for generating too-short (or long) sequences. The $\lambda$'s are hyperparameters to be tuned using on a small amount of held-out development data. 

\section{Experiments}
\label{sec:experiments}
We evaluate our model on three natural language processing tasks, abstractive sentence summarisation, machine translation and morphological inflection generation. For each task, we compare the performance of the direct model, noisy channel model, and the interpolation of the two models.

\subsection{Abstractive Sentence Summarisation}
Sentence summarisation is the problem of constructing a shortened version of a sentence while preserving the majority of its meaning. In contrast to extractive summarisation, which can only copy words from the original sentence, abstractive summarisation permits arbitrary rewording of the sentence.
The dataset \citep{DBLP:conf/emnlp/RushCW15} that we use is constructed by pairing the first sentence and the headline of each article from the annotated Gigaword corpus \citep{graff2003english,napoles2012annotated}. There are 3.8m, 190k and 381k sentence pairs in the training, validation and test sets, respectively. \cite{yu:2016} filtered the dataset by restricting the lengths of the input and output sentences to be no greater than 50 and 25 tokens, respectively. From the filtered data, they further sampled 1 million sentence pairs for training. We experimented on training the direct model and channel model with both the sampled 1 million and the full 3.8 million parallel data. The language model is trained on the target side of the parallel data, i.e. the headlines. We evaluated the generated summaries of 2000 randomly sampled sentence pairs using full length ROUGE F1. This setup is in line with the previous work on this task \citep{DBLP:conf/emnlp/RushCW15,chopra,gulcehre2016pointing,yu:2016}.

The same configuration is used to train the direct model and the channel model. The loss (Equation \ref{loss}) is optimized by Adam \citep{DBLP:journals/corr/KingmaB14}, with initial learning rate of 0.001. We use LSTMs with 1 layer for both the encoder and decoders, with hidden units of 256. The mini-batch size is 32, and dropout of 0.2 is applied to the input and output of LSTMs. For the language model, we use a 2-layer LSTM with 1024 hidden units and 0.5 dropout. The learning rate is 0.0001. All the hyperparameters are optimised via grid search on the perplexity of the validation set. During decoding, beam search is employed with the number of proposals generated by the direct model $K_1 = 20$, and the number of best candidates selected by the noisy channel model $K_2 = 10$.

\begin{table}[t]\centering
	\begin{tabular}{@{}lccccc@{}}
		\toprule
		Model & \# Parallel data & \# Data for LM & RG-1 & RG-2  & RG-L \\
		\midrule
		direct (uni)$^*$ & 1.0m & - & 30.94 & 14.20 & 28.72 \\
		%direct (bi) & 1.0m & - & 30.78 & 14.67 & 28.57 \\
		direct (bi) & 1.0m & - & 31.25 & 14.52 & 29.03 \\
		direct (bi) & 3.8m & - & 33.82 & 16.66 & 31.50 \\
		\midrule
		channel + LM + bias (uni)$^*$	& 1.0m & 1.0m & 31.92 & 14.75 & 29.58 \\
		channel + LM + bias (bi)	& 1.0m & 1.0m & 31.96 & 14.89 & 29.51 \\
		direct + channel + LM + bias (uni) & 1.0m & 1.0m & 33.07 & 15.21 & 30.29 \\
		direct + channel + LM + bias (bi) & 1.0m & 1.0m & 33.18 & 15.65 & 30.45 \\
		channel + LM + bias (uni)$^*$ & 1.0m & 3.8m & 32.59 & 15.05 & 30.06 \\
		channel + LM + bias (bi) & 1.0m & 3.8m & 32.65 & 14.95 & 30.23 \\
		%direct + LM + bias (bi) & 1.0m & 3.8m & 30.14 & 14.12 & 28.10 \\
		direct + LM + bias (bi) & 1.0m & 3.8m & 31.25 & 14.52 & 29.03 \\
		direct + channel + LM + bias (uni) & 1.0m & 3.8m  & 33.16 & 15.63 & 30.53 \\
		direct + channel + LM + bias (bi) & 1.0m & 3.8m  & 33.21  & 15.65 & 30.60 \\
		chanel + LM + bias (bi) & 3.8m & 3.8m & 34.12 & 16.41 & 31.38\\
		%direct + LM + bias (bi) & 3.8m & 3.8m & 31.93 & 16.05 & 29.94 \\
		direct + LM + bias (bi) & 3.8m & 3.8m & 33.82 & 16.66 & 31.50 \\
		direct + channel + LM + bias (bi) & 3.8m & 3.8m & \bfseries{34.41} & \bfseries{16.86} & \bfseries{31.83} \\
		\bottomrule
	\end{tabular}
	\caption {ROUGE F1 scores on the sentence summarisation test set. The `uni' and `bi' in the parentheses denote the encoder for the model proposing candidates is a unidirectional LSTM or bidirectional LSTM. Those rows marked with an $*$ denote models that process their input online.}
	\label{test_rg}
\end{table}

\begin{table}[t]\centering
	\begin{tabular}{@{}lccccc@{}}
		\toprule
		Model & \# Parallel data & \# Unpaired data & RG-1 & RG-2  & RG-L \\
		\midrule
		ABS+ & 3.8m & - & 29.55 & 11.32 & 26.42 \\
		RAS-LSTM & 3.8m & - & 32.55 & 14.70 & 30.03 \\
		RAS-Elman & 3.8m & - & 33.78 & 15.97 & 31.15 \\
		Pointing unkown words & 3.8m & - & \bfseries{35.19} & 16.66 & \bfseries{32.51} \\
		ASC + FSC & 1.0m & 3.8m & 31.09 & 12.79 & 28.97 \\
		ASC + FSC & 3.8m & 3.8m & 34.17 & 15.94 & 31.92 \\
		\midrule
		direct + channel + LM + bias (bi) & 1.0m & 3.8m & 33.21  & 15.65 & 30.60 \\
		direct + channel + LM + bias (bi) & 3.8m & 3.8m & 34.41 & \bfseries{16.86} & 31.83 \\
		\bottomrule 
	\end{tabular}
	\caption {Overview of results on the abstractive sentence summarisation task. ABS+ \citep{DBLP:conf/emnlp/RushCW15} is the attentive model with bag-of-words as the encoder. RAS-LSTM and RAS-Elman \citep{chopra} are the sequence to sequence models with attention with the RNN cell implemented as LSTMs and an Elman architecture \citep{elman1990finding}, respectively. Pointing the unknown words \citep{gulcehre2016pointing} uses pointer networks \citep{vinyals2015pointer} to select the output token from the input sequence in order to avoid generating unknown tokens. ASC + FSC \citep{miao2016} is the semi-supervised model based on a variational autoencoder. } 
	\label{prev_work}
\end{table}

Table \ref{test_rg} presents the ROUGE-F1 scores of the test set from the direct model, noisy channel model (channel + LM + bias), the interpolation of the direct model and the noisy channel model (direct + channel + LM + bias), and the interpolation of the direct model and language model (direct + LM + bias) trained on different sizes of data. The noisy channel model with the language model trained on the target side of the 1 million parallel data outperforms the direct model by approximately 1 point. Such improvement indicates that the language model helps improve the quality of the output sequence when no extra unlabelled data is available. Training the language model with all the headlines in the dataset, i.e. 3.8 million sentences, gives a further boost to the ROUGE score. This is in line with our expectation that the model benefits from adding large amounts of unlabelled data. The interpolation of the direct model, channel model, language model and bias of the output length achieves the best results --- the ROUGE score is close to the direct model trained on all the parallel data. Although there is still improvement, when the direct model is trained with more data, the gap between the direct model and the noisy channel model is smaller. No gains is observed if the language model is combined with the direct model. We find that as we increase the weight of the language model, the result is getting worse.

Table \ref{prev_work} surveys published results on this task, and places our best models in the context of the current state-of-the-art results. ABS+ \citep{DBLP:conf/emnlp/RushCW15}, RAS-LSTM and RAS-Elman \citep{chopra} are different variations of the attentive models. {\it Pointing the unkown words} uses pointer networks \citep{vinyals2015pointer} to select the output token from the input sequence in order to avoid generating unknown tokens. ASC + FSC \citep{miao2016} is a semi-supervised model based on a variational autoencoder. Trained on 1m paired samples and 3.8m unpaired samples, the noisy channel achieves comparable or better results than (direct) models trained with 3.8m paired samples. Compared to \cite{miao2016}, whose ASC + FSC models is an alternative strategy for using unpaired data, the noisy channel is significantly more effective --- 33.21 versus 31.09 in ROUGE-1.

Finally, motivated by the qualitative observation that noisy channel model outputs were quite fluent and often used reformulations of the input rather than a strict compression (which would be poorly scored by  ROUGE), we carried out a human preference evaluation whose results are summarised in Table~\ref{tab:human}. This confirms that noisy channel summaries are strongly preferred over those of the direct model.

\begin{table}[h]\centering
	\begin{tabular}{lc}
		\toprule
		Model & count \\
		\midrule
both bad &188 \\ 
both good &106 \\
direct $>$ noisy channel &135 \\
noisy channel $>$ direct & \bfseries{212} \\
\bottomrule 
	\end{tabular}
	\caption{Preference ratings for 641 segments from the test set (each segment had ratings from at least 2 raters with $\ge$ 50\% agreement on the label and where one label had a plurality of the votes).} 
	\label{tab:human}
\end{table}

\subsection{Machine Translation}
We next evaluate our models on a Chinese--English machine translation task. We used parallel data with 184k sentence pairs (from the FBIS corpus, LDC2003E14) and monolingual data with 4.3 million of English sentences (selected from the English Gigaword).
The training data is preprocessed by lowercasing the English sentences, replacing digits with `\#' token, and replacing tokens appearing less than 5 times with an UNK token. This results in vocabulary sizes of 30k and 20k for Chinese sentences and English sentences, respectively.

The models are trained using Adam \citep{DBLP:journals/corr/KingmaB14} with initial learning rate of 0.001 for the direct model and the channel model, and 0.0001 for the language model. The LSTMs for the direct and channel models have 512 hidden units and 1 layer, and 2 layers with 1024 hidden units per layer for the language model. Dropout of 0.5 on the input and output of LSTMs is set for all the model training. The noisy channel decoding uses $K_1$ = 20 and $K_2$ = 10 as the beam sizes.

Table \ref{mt-result} lists the translation performance of different models in BLEU scores. To set benchmarks, we train the vanilla and attentional sequence to sequence models \citep{sutskever:2014,bahdanau:2015} using the same parallel data. For direct models, we leverage bidirectional LSTMs as the encoder for this task. We can see that the vanilla sequence to sequence model behaves poorly due to the small amounts of parallel data. By contrast, the direct model (SSNT) and the attentional model work relatively well, with the attentional model outperforming the SSNT direct model. Although these models both directly model $p(\boldsymbol{y} \mid \boldsymbol{x})$, this result is unsurprising because the SSNT direct model is most effective when the alignment between sequences is largely monotonic, and Chinese--English translation word orders diverge considerably. However, despite this limitation, the noisy channel model is approximately 3 points higher in BLEU than the direct model, and the combination of noisy channel and direct model gives extra boost. Confirming the empirical findings of prior work (and in line with theoretical predictions), the interpolation of the direct model and language model is not effective.

\begin{table}[t]\centering
	\begin{tabular}{@{}lc@{}}
		\toprule
		Model & BLEU \\
		\midrule
		seq2seq w/o attention & 11.19 \\
		seq2seq w/ attention & 25.27\\
		direct (bi) & 23.33 \\
		\midrule
		direct + LM + bias (bi) & 23.33 \\
		channel + LM + bias (bi) & 26.28 \\
		direct + channel + LM + bias (bi) & \bfseries{26.44} \\
		\bottomrule
	\end{tabular}
	\caption {BLEU scores from different models for the Chinese to English machine translation task.}
	\label{mt-result}
\end{table}

\subsection{Morphological Inflection Generation}
Morphological inflection is the task of generating a target (inflected form) word from a source word (base form), given a morphological attribute, e.g. number, tense, and person etc.. It is useful for reducing data sparsity issues in translating morphologically rich languages. The transformation from the base form to the inflected form is usually to add prefix or suffix, or to do character replacement. The dataset \citep{DBLP:conf/naacl/DurrettD13} that we use in the experiments is created from Wiktionary, including inflections for German nouns, German verbs, Spanish Verbs, Finnish noun and adjective, and Finnish verbs. We only experimented on German nouns and German verbs, as German nouns is the most difficult task\footnote{While state-of-the-art systems can achieve 99\% accuracies on Spanish verbs and Finnish verbs, they can only get 89\% accuracy on German nouns.}, and the direct model does not perform as well as other state-of-the-art systems on German verbs. The train/dev/test split for German nouns is 2364/200/200, and for German verbs is 1617/200/200. There are 8 and 27 inflection types in German nouns and German verbs, respectively. Following previous work, we learn a separate model for each type of inflection independent
of the other inflections. We report results on the average accuracy across different inflections.
Our language models were trained on word types extracted by running a morphological analysis tool on the WMT 2016 monolingual data and extracting examples of appropriately inflected word forms.\footnote{\url{http://www.statmt.org/wmt16/translation-task.html}}
After annotation the number of instances for training the language model ranged from 300k to 3.8m for different inflection types in German nouns, and from 200 to 54k in German verbs.

The experimental setup that we use on this task is $K_1$ = 60, $K_2$ = 30,
\begin{itemize}
    \item direct and channel model: 1 layer LSTM with 128 hidden, $\eta = 0.001$, dropout = 0.5.
    \item language model: 2 layer LSTM with 512 hidden, $\eta = 0.0001$, dropout = 0.5.
\end{itemize}

Table \ref{morph-result} summarises the results from our models. On both datasets, the noisy channel model (channel + LM + bias) does not perform as well as the direct model, but the interpolation of the direct model and noisy channel model (direct + channel + LM + bias) significantly outperforms the direct model. The interpolation of the direct model and language model (direct + LM + bias) achieves better results than the direct model and the noisy channel model on German nouns, but not on German verbs. For further comparison, we also included the state-of-the-art results as benchmarks. NCK15 \citep{DBLP:conf/naacl/NicolaiCK15} tackles the task based on the three-stage approach: (1) align the source and target word, (2) extract inflection rules, (3) apply the rule to new examples. FTND16 \citep{DBLP:conf/naacl/FaruquiTND16} is based on neural sequence to sequence models. Both models (NCK15+ and FTND16+) rerank the candidate outputs by the scores predicted from n-gram language models, together with other features. 

\begin{figure}
		\begin{subfigure}{.5\textwidth}
			\centering
			\begin{tabular}{@{}lc@{}}
		    \toprule
		    Model & Acc. \\
		    \midrule
		    NCK15 & 88.60\\
		    FTND16 & 88.12 \\
		    NCK15+ & 89.90\\
		    FTND16+ & 89.31\\
		    \midrule
		    direct (uni) & 82.25 \\
		    direct (bi) & 87.68 \\
		    \midrule
		    channel + LM + bias (uni) & 78.38\\
		    channel + LM + bias (bi) & 78.13 \\
		    direct + LM + bias (bi) & 90.31 \\
		    direct + channel + LM + bias (uni) & 88.44 \\
		    direct + channel + LM + bias (bi) & \bfseries{90.94} \\ 
		    \bottomrule
	\end{tabular}
			\caption{}
			\label{fig:sfig1}
		\end{subfigure}%
		\begin{subfigure}{.5\textwidth}
			\centering
			\begin{tabular}{@{}lc@{}}
				\toprule
				Model & Acc. \\
				\midrule
				NCK15 & 97.50 \\
				FTND16 & \bfseries{97.92} \\
				NCK15+ & 97.90 \\
				FTND16+ & 97.11 \\
				\midrule
				direct (uni) & 87.85\\
				direct (bi) & 94.83\\
				\midrule
				channel + LM + bias (uni) & 84.42\\
				channel + LM + bias (bi) & 92.13\\
				direct + LM + bias (bi) & 94.83 \\
				direct + channel + LM + bias (uni) & 92.20\\
				direct + channel + LM + bias (bi) & 97.15\\ 
				\bottomrule
			\end{tabular}
			\caption{}
			\label{fig:sfig2}
		\end{subfigure}
		\caption{Accuracy on morphological inflection of German nouns (a), and German verbs (b). NCK15 \citep{DBLP:conf/naacl/NicolaiCK15} and FTND16 \citep{DBLP:conf/naacl/FaruquiTND16} are previous state-of-the-art on this task, with NCK15 based on feature engineering, and FTND16 based on neural networks. NCK15+ and FTND16+ are the semi-supervised setups of these models.}
		\label{morph-result}
\end{figure}

\section{Analysis}
By observing the output generated by the direct model and noisy channel model, we find (in line with theoretical critiques of conditional models) that the direct model may leave out key information. By contrast, the noisy channel model does seem to avoid this issue. To illustrate, in Example~1 (see Appendix~B) in Table~\ref{example}, the direct model ignores the key phrase `coping with', resulting in incomplete meaning, but the noisy channel model covers it. Similarly, in Example 6, the direct model does not translate the Chinese word corresponding to `investigation'. We also observe that while the direct model mostly copies words from the source sentence, the noisy channel model prefers generating paraphrases. For instance, in Example 2, while the direct model copies the word `accelerate' in the generated output, the noisy channel model generate `speed up' instead. While one might argue that copying is a preferable compression technique than paraphrasing (as long as it produces grammatical outputs), it does show the power of these models. 

\section{Related work}
Noisy channel decompositions have been successfully used in a variety of problems, including speech recognition~\citep{jelinek:1998}, machine translation~\citep{brown:1993}, spelling correction~\citep{brill:2000}, and question answering~\citep{echihabi:2003}. The idea of adding language models and monolingual data in machine translation has been explored in earlier work. \cite{gulcehre:2015} propose two strategies of combining a language model with a neural sequence to sequence model. In shallow fusion, during decoding the sequence to sequence model (direct model) proposes candidate outputs and these candidates are reranked based on the scores calculated by a weighted sum of the probability of the translation model and that of the language model. In deep fusion, the language model is integrated into the decoder of the sequence to sequence model by concatenating their hidden state at each time step. \cite{sennrich:2016} incorporate target language unpaired training data by doing back-translation to create synthetic parallel training data. While this technique is quite effective, its practicality seems limited to problems where the inputs and outputs contain roughly the same information (such as translation). \cite{cheng:2016} leverages the abundant monolingual data by doing multitask learning with an autoencoding objective.

A number of papers have remarked on the tendency for content to get dropped (or repeated) in translation. \citet{liu:2016} propose translating in both a left-to-right and a left-to-right direction and seeking a consensus. \citet{tu:2016} propose augmenting a direct model's decoding objective with a reverse translation model (similar to our channel model except it conditions on the direct model's output RNN's hidden states rather than the words); however, that work just reranks complete translation hypotheses rather than developing a model that permits an incremental search.

Another trend of work that is related to our model is the investigation of making online prediction for machine translation \citep{gu:2016,grissom:2014,sankaran:2010} and speech recognition \citep{hwang:2016,jaitly2015neural}.

Our direct model (and channel model) shares the idea of introducing stochastic latent variables to neural networks with several papers and marginalising these during training. Examples include connectionist
temporal classification (CTC) \citep{graves2006connectionist} and the more recent segmental recurrent neural networks (SRNN) \citep{kong2015segmental}. Compared to these models, our direct model has the advantage of capturing unbounded dependencies of output words. The direct model is closely related to the sequence transduction model \citep{graves2012sequence} in the way of modeling the probability of predicting output tokens and marginalizing latent variables using dynamic programming. However, rather than modeling the joint distribution over outputs and alignments by inserting null symbols
into the output sequence, our direct model defines a separate latent alignment variable, with alignment distribution defined with neural networks. Similar to our work, the model in \citep{alkhoulialignment} is decomposed into the alignment model and the model of word predictions. The two models are trained separately and combined during decoding, with subsequent refinements using a Viterbi-EM approximation. By contrast, in our direct and channel models, the latent and observed components of the models are trained jointly using a dynamic program to exactly marginalise the unobserved variables.

\section{Conclusion}

We have presented and empirically validated a noisy channel transduction model that uses component models based on recurrent neural networks. This formulation lets us use unpaired outputs to estimate the parameters of the source model and input-output pairs to train the channel model. Despite the channel model's ability to condition on long sequences, we are able to maintain tractable decoding by using a latent segmentation variable that breaks the conditioning context up into a series of monotonically growing segments. Our experiments show that this model makes excellent use of unpaired training data.

\bibliography{biblio}

\begin{thebibliography}{39}
\providecommand{\natexlab}[1]{#1}
\providecommand{\url}[1]{\texttt{#1}}
\expandafter\ifx\csname urlstyle\endcsname\relax
  \providecommand{\doi}[1]{doi: #1}\else
  \providecommand{\doi}{doi: \begingroup \urlstyle{rm}\Url}\fi

\bibitem[Alkhouli et~al.(2016)Alkhouli, Bretschner, Peter, Hethnawi, Guta, and
  Ney]{alkhoulialignment}
Tamer Alkhouli, Gabriel Bretschner, Jan-Thorsten Peter, Mohammed Hethnawi,
  Andreas Guta, and Hermann Ney.
\newblock Alignment-based neural machine translation.
\newblock In \emph{Proc. Machine Translation}, 2016.

\bibitem[Bahdanau et~al.(2015)Bahdanau, Cho, and Bengio]{bahdanau:2015}
Dzmitry Bahdanau, Kyunghyun Cho, and Yoshua Bengio.
\newblock Neural machine translation by jointly learning to align and
  translate.
\newblock In \emph{Proc. ICLR}, 2015.

\bibitem[Brill \& Moore(2000)Brill and Moore]{brill:2000}
Eric Brill and Robert~C. Moore.
\newblock An improved error model for noisy channel spelling correction.
\newblock In \emph{Proc. ACL}, 2000.

\bibitem[Brown et~al.(1993)Brown, Pietra, Pietra, and Mercer]{brown:1993}
Peter~F. Brown, Stephen A.~Della Pietra, Vincent J.~Della Pietra, and
  Robert.~L. Mercer.
\newblock The mathematics of statistical machine translation: Parameter
  estimation.
\newblock \emph{Computational Linguistics}, 19:\penalty0 263--311, 1993.

\bibitem[Cheng et~al.(2016)Cheng, Xu, He, He, Wu, Sun, and Liu]{cheng:2016}
Yong Cheng, Wei Xu, Zhongjun He, Wei He, Hua Wu, Maosong Sun, and Yang Liu.
\newblock Semi-supervised learning for neural machine translation.
\newblock In \emph{Proc. ACL}, 2016.

\bibitem[Chopra et~al.(2016)Chopra, Auli, and Rush]{chopra}
Sumit Chopra, Michael Auli, and Alexander~M. Rush.
\newblock Abstractive sentence summarization with attentive recurrent neural
  networks.
\newblock In \emph{Proc. NAACL}, 2016.

\bibitem[Durrett \& DeNero(2013)Durrett and DeNero]{DBLP:conf/naacl/DurrettD13}
Greg Durrett and John DeNero.
\newblock Supervised learning of complete morphological paradigms.
\newblock In \emph{HLT-NAACL}, 2013.

\bibitem[Echihabi \& Marcu(2003)Echihabi and Marcu]{echihabi:2003}
Abdessamad Echihabi and Daniel Marcu.
\newblock A noisy-channel approach to question answering.
\newblock In \emph{Proc. ACL}, 2003.

\bibitem[Elman(1990)]{elman1990finding}
Jeffrey~L. Elman.
\newblock Finding structure in time.
\newblock \emph{Cognitive science}, 14\penalty0 (2):\penalty0 179--211, 1990.

\bibitem[Faruqui et~al.(2016)Faruqui, Tsvetkov, Neubig, and
  Dyer]{DBLP:conf/naacl/FaruquiTND16}
Manaal Faruqui, Yulia Tsvetkov, Graham Neubig, and Chris Dyer.
\newblock Morphological inflection generation using character sequence to
  sequence learning.
\newblock In \emph{Proc. HLT-NAACL}, 2016.

\bibitem[Graff et~al.(2003)Graff, Kong, Chen, and Maeda]{graff2003english}
David Graff, Junbo Kong, Ke~Chen, and Kazuaki Maeda.
\newblock English gigaword.
\newblock \emph{Linguistic Data Consortium, Philadelphia}, 2003.

\bibitem[Graves(2012)]{graves2012sequence}
Alex Graves.
\newblock Sequence transduction with recurrent neural networks.
\newblock \emph{arXiv preprint arXiv:1211.3711}, 2012.

\bibitem[Graves et~al.(2006)Graves, Fern{\'a}ndez, Gomez, and
  Schmidhuber]{graves2006connectionist}
Alex Graves, Santiago Fern{\'a}ndez, Faustino Gomez, and J{\"u}rgen
  Schmidhuber.
\newblock Connectionist temporal classification: labelling unsegmented sequence
  data with recurrent neural networks.
\newblock In \emph{ICML}, 2006.

\bibitem[{Grissom II} et~al.(2014){Grissom II}, Boyd-Graber, He, Morgan, and
  {Daum\'{e} III}]{grissom:2014}
Alvin {Grissom II}, Jordan Boyd-Graber, He~He, John Morgan, and Hal {Daum\'{e}
  III}.
\newblock Don't until the final verb wait: Reinforcement learning for
  simultaneous machine translation.
\newblock In \emph{Empirical Methods in Natural Language Processing}, 2014.

\bibitem[Gu et~al.(2016)Gu, Neubig, Cho, and Li]{gu:2016}
Jiatao Gu, Graham Neubig, Kyunghyun Cho, and Victor~O.K. Li.
\newblock Learning to translate in real-time with neural machine translation.
\newblock \emph{CoRR}, abs/1610.00388, 2016.

\bibitem[G{\"{u}}l{\c{c}}ehre et~al.(2015)G{\"{u}}l{\c{c}}ehre, Firat, Xu, Cho,
  Barrault, Lin, Bougares, Schwenk, and Bengio]{gulcehre:2015}
{\c{C}}aglar G{\"{u}}l{\c{c}}ehre, Orhan Firat, Kelvin Xu, Kyunghyun Cho,
  Lo{\"{\i}}c Barrault, Huei{-}Chi Lin, Fethi Bougares, Holger Schwenk, and
  Yoshua Bengio.
\newblock On using monolingual corpora in neural machine translation.
\newblock \emph{CoRR}, abs/1503.03535, 2015.

\bibitem[G{\"{u}}l{\c{c}}ehre et~al.(2016)G{\"{u}}l{\c{c}}ehre, Ahn, Nallapati,
  Zhou, and Bengio]{gulcehre2016pointing}
{\c{C}}aglar G{\"{u}}l{\c{c}}ehre, Sungjin Ahn, Ramesh Nallapati, Bowen Zhou,
  and Yoshua Bengio.
\newblock Pointing the unknown words.
\newblock \emph{CoRR}, abs/1603.08148, 2016.

\bibitem[Hochreiter \& Schmidhuber(1997)Hochreiter and
  Schmidhuber]{hochreiter1997long}
Sepp Hochreiter and J{\"u}rgen Schmidhuber.
\newblock Long short-term memory.
\newblock \emph{Neural computation}, 9\penalty0 (8):\penalty0 1735--1780, 1997.

\bibitem[Hwang \& Sung(2016)Hwang and Sung]{hwang:2016}
Kyuyeon Hwang and Wonyong Sung.
\newblock Character-level incremental speech recognition with recurrent neural
  networks.
\newblock In \emph{Proc. ICASSP}, 2016.

\bibitem[Jaitly et~al.(2016)Jaitly, Sussillo, Le, Vinyals, Sutskever, and
  Bengio]{jaitly2015neural}
Navdeep Jaitly, David Sussillo, Quoc~V Le, Oriol Vinyals, Ilya Sutskever, and
  Samy Bengio.
\newblock A neural transducer.
\newblock \emph{Proc. NIPS}, 2016.

\bibitem[Jelinek(1998)]{jelinek:1998}
Frederick Jelinek.
\newblock \emph{Statistical Methods for Speech Recognition}.
\newblock MIT, 1998.

\bibitem[Kalchbrenner \& Blunsom(2013)Kalchbrenner and
  Blunsom]{kalchbrenner:2013}
Nal Kalchbrenner and Phil Blunsom.
\newblock Recurrent continuous translation models.
\newblock In \emph{Proc. EMNLP}, 2013.

\bibitem[Kingma \& Ba(2015)Kingma and Ba]{DBLP:journals/corr/KingmaB14}
Diederik~P. Kingma and Jimmy Ba.
\newblock Adam: {A} method for stochastic optimization.
\newblock In \emph{Proc. ICIR}, 2015.

\bibitem[Klein \& Manning(2001)Klein and Manning]{klein:2001}
Dan Klein and Christopher~D. Manning.
\newblock Conditional structure versus conditional estimation in nlp models.
\newblock In \emph{Proc. EMNLP}, 2001.

\bibitem[Kong et~al.(2016)Kong, Dyer, and Smith]{kong2015segmental}
Lingpeng Kong, Chris Dyer, and Noah~A Smith.
\newblock Segmental recurrent neural networks.
\newblock \emph{Proc. ICLR}, 2016.

\bibitem[Liu et~al.(2016)Liu, Utiyama, Finch, and Sumita]{liu:2016}
Lemao Liu, Masao Utiyama, Andrew Finch, and Eiichiro Sumita.
\newblock Agreement on target-bidirectional neural machine translation.
\newblock In \emph{Proc. NAACL}, 2016.

\bibitem[Miao \& Blunsom(2016)Miao and Blunsom]{miao2016}
Yishu Miao and Phil Blunsom.
\newblock Language as a latent variable: Discrete generative models for
  sentence compression.
\newblock In \emph{Proc. EMNLP}, 2016.

\bibitem[Napoles et~al.(2012)Napoles, Gormley, and
  Van~Durme]{napoles2012annotated}
Courtney Napoles, Matthew Gormley, and Benjamin Van~Durme.
\newblock Annotated gigaword.
\newblock In \emph{Proceedings of the Joint Workshop on Automatic Knowledge
  Base Construction and Web-scale Knowledge Extraction}, 2012.

\bibitem[Nicolai et~al.(2015)Nicolai, Cherry, and
  Kondrak]{DBLP:conf/naacl/NicolaiCK15}
Garrett Nicolai, Colin Cherry, and Grzegorz Kondrak.
\newblock Inflection generation as discriminative string transduction.
\newblock In \emph{Proc. HLT-NAACL}, 2015.

\bibitem[Rush et~al.(2015)Rush, Chopra, and Weston]{DBLP:conf/emnlp/RushCW15}
Alexander~M. Rush, Sumit Chopra, and Jason Weston.
\newblock A neural attention model for abstractive sentence summarization.
\newblock In \emph{Proc. EMNLP}, 2015.

\bibitem[Sankaran et~al.(2010)Sankaran, Grewal, and Sarkar]{sankaran:2010}
Baskaran Sankaran, Ajeet Grewal, and Anoop Sarkar.
\newblock Incremental decoding for phrase-based statistical machine
  translation.
\newblock In \emph{Proc. WMT}, 2010.

\bibitem[Sennrich et~al.(2016)Sennrich, Haddow, and Birch]{sennrich:2016}
Rico Sennrich, Barry Haddow, and Alexandra Birch.
\newblock Improving neural machine translation models with monolingual data.
\newblock In \emph{Proc. ACL}, 2016.

\bibitem[Shannon(1948)]{shannon:1948}
Claude Shannon.
\newblock A mathematical theory of communication.
\newblock \emph{Bell System Technical Journal}, 27\penalty0 (3):\penalty0
  379--423, 1948.

\bibitem[Sima'an(1996)]{simaan:1996}
Khalil Sima'an.
\newblock Computational complexity of probabilistic disambiguation by means of
  tree-grammars.
\newblock In \emph{Proc. COLING}, 1996.

\bibitem[Sutskever et~al.(2014)Sutskever, Vinyals, and Le]{sutskever:2014}
Ilya Sutskever, Oriol Vinyals, and Quoc~V. Le.
\newblock Sequence to sequence learning with neural networks.
\newblock In \emph{Proc. NIPS}, 2014.

\bibitem[Tillmann et~al.(1997)Tillmann, Vogel, Ney, and
  Zubiaga]{tillmann1997dp}
Christoph Tillmann, Stephan Vogel, Hermann Ney, and Alex Zubiaga.
\newblock A {DP}-based search using monotone alignments in statistical
  translation.
\newblock In \emph{Proc. EACL}, 1997.

\bibitem[Tu et~al.(2016)Tu, Liu, Shang, Liu, and Li]{tu:2016}
Zhaopeng Tu, Yang Liu, Lifeng Shang, Xiaohua Liu, and Hang Li.
\newblock Neural machine translation with reconstruction.
\newblock \emph{CoRR}, abs/1611.01874, 2016.

\bibitem[Vinyals et~al.(2015)Vinyals, Fortunato, and
  Jaitly]{vinyals2015pointer}
Oriol Vinyals, Meire Fortunato, and Navdeep Jaitly.
\newblock Pointer networks.
\newblock In \emph{Proc. NIPS}, 2015.

\bibitem[Yu et~al.(2016)Yu, Buys, and Blunsom]{yu:2016}
Lei Yu, Jan Buys, and Phil Blunsom.
\newblock Online segment to segment neural transduction.
\newblock In \emph{Proc. EMNLP}, 2016.

\end{thebibliography}
\bibliographystyle{iclr2017_conference}

\newpage

\appendix

\section{Algorithm}
\label{sec:algo_appendix}

\begin{algorithm*}[ht]                      
	\caption{Noisy Channel Decoding}        
	\label{decode} 
	%\etg{The notation $\topk(K_1)_{y \in \mathcal{V}}$ is hard to parse in the lines below, and could be cleaned up, especially when followed by $p(...)$ without a space. The semantics of $\topk(K)_{\alpha \in \mathcal{A}}$ must be explained somewhere.}
	\begin{algorithmic}          
        \State \textbf{Notation: } $Q$ is the Viterbi matrix, bp is the backpointer, $W$ stores the predicted tokens, $\mathcal{V}$ refers to the vocabulary, $I=|\boldsymbol{x}|$, and $J_\text{max}$ denotes the maximum number of output tokens that can be predicted.
		\State \textbf{Input: } source sequence $\boldsymbol{x}$
		\State \textbf{Output: } best output sequence $\boldsymbol{y^*}$
		\State \textbf{Initialisation: } $Q \in \mathbb{R}^{I \times J_\text{max}\times K_1}$, bp $\in \mathbb{N}^{I \times J_\text{max}\times K_1}$,  $W \in \mathbb{N}^{I \times J_\text{max}\times K_1}$,
		
		\ \ \ \ \ \ \ \ \ \ \ \ \ \ \ \ \ \
		$Q_{temp} \in \mathbb{R}^{K_1}$, $bp_{temp} \in \mathbb{N}^{K_1}$, $W_{temp} \in \mathbb{N}^{K_1}$
		\For{$i \in [1, I]$}
			\State $Q_{temp} \gets \topk(K_1)_{y \in \mathcal{V}}q(z_1 = i) $ $q(y\ |\ \textsc{start}, z_1, \boldsymbol{x}_{1}^{z_1})$ \Comment Candidates generated by $q(\boldsymbol{y}\ |\ \boldsymbol{x})$.
			\State $bp_{temp}\gets 0$
			\State $W_{temp} \gets \argtopk(K_1)_{y \in \mathcal{V}}q(z_1 = i)$ $q(y\ |\ \textsc{start}, z_1, \boldsymbol{x}_1^{z_1})$
			
			\State $Q[i, 1] \gets \topk(K_2)_{y \in W_{temp}} O_{\boldsymbol{x}_1^i, y}$ \Comment Rerank the candidates by objective ($O$).
			\State $W[i,1] \gets \argtopk(K_2)_{y \in W_{temp}}O_{\boldsymbol{x}_1^i, y}$
		\EndFor
		\For{$j\in[2, J_\text{max}]$}
			\For{$i \in [1, I]$}
				\State $Q_{temp} \gets \topk(K_1)_{y \in \mathcal{V}, k \in [1, i]} Q[k,j-1] \cdot$ $q(z_j = i\ |\ z_{j-1} = k)q(y\ |\ \boldsymbol{y}_1^{j-1}, z_j, \boldsymbol{x}_1^{z_j})$
				\State $bp_{temp} , W_{temp} \gets \argtopk(K_1)_{y \in \mathcal{V}, k \in [1, i]} $ $Q[k,j-1]q(z_j = i\ |\ z_{j-1} = k) \cdot$
				
				\ \ \ \ \ \ \ \ \ \ \ \ \ \ \ \ \ \ \ \ \ \ \ \ \ \ \ \ \ \ \ \ \ \ \ \ \ \ \ \ \ \ \ \ \ \ \ \ \ \ \ \ \ \ \ \ \ \ \ \ \ \ \ \ \ \ \ \ \ \ \ \ \ \ \ \ \ \ \ \ \ \ \ \ \ \ \ \ \ \ \ \ \ \ \ \ \ \ \ \ \ \ \ \ \ \ \ \ \ \ \ \ \ \ $q(y\ |\ \boldsymbol{y}_1^{j-1}, z_j, \boldsymbol{x})$
				
				\State $Y \gets \candidate(bp_{temp}, W_{temp})$ \Comment Get partial candidate $\boldsymbol{y}_1^j$.
				%\State $Q[i,j] \gets \topk(K_2)_{\boldsymbol{y}_j \in Y}$ $p(\boldsymbol{x}_1^i\ |\ \boldsymbol{y}_1^j) p(\boldsymbol{y}_1^j)$
				\State $Q[i,j] \gets \topk(K_2)_{\boldsymbol{y}_j \in Y} O_{\boldsymbol{x}_1^i, \boldsymbol{y}_1^j}$ 
				\State $bp[i,j] , W[i,j] \gets %\argtopk(K_2)_{\boldsymbol{y}_1^j \in Y}$ $p(\boldsymbol{x}_1^i\ |\ \boldsymbol{y}_1^j) p(\boldsymbol{y}_1^j)$
				\argtopk(K_2)_{\boldsymbol{y}_1^j \in Y}$ $O_{\boldsymbol{x}_1^i, \boldsymbol{y}_1^j}$
			\EndFor
		\EndFor 
		\State
		\Return a sequence of words stored in $W$ by following backpointers starting from $(I,\argmax_j Q[I, j])$.
	\end{algorithmic}
\end{algorithm*}

\section{Example outputs}
\label{sec:outputs}

\begin{table}
    \begin{tabular}{@{}lp{12cm}@{}}
		\toprule
		 &\bfseries{Summarisation} \\
		\bfseries{Example 1:} & \\
		\bfseries{source:} & the european commission on health and consumers protection $-$lrb$-$ \_unk\_ $-$rrb$-$ has offered cooperation to indonesia in coping with the spread of avian influenza in the country , official news agency antara said wednesday . \\
		\bfseries{reference:} & eu offers indonesia cooperation in \underline{avian flu eradication} \\
		\bfseries{direct:} & eu offers cooperation to indonesia in \underline{avian flu} \\
		\bfseries{nc:} & eu offers cooperation to indonesia in \underline{coping with bird flu} \\
		\midrule
		\bfseries{Example 2:} & \\
        \bfseries{source:} & vietnam will \underline{accelerate} the export of industrial goods mainly by developing auxiliary industries , and helping enterprises sharpen competitive edges , according to the ministry of industry on thursday . \\
		\bfseries{reference:} & vietnam to \underline{boost} industrial goods export \\
		\bfseries{direct:} & vietnam to \underline{accelerate} export of industrial goods \\
		\bfseries{nc:} & vietnam to \underline{speed up} export of industrial goods \\
		\midrule
		\bfseries{Example 3:} & \\
        \bfseries{source:} & japan 's toyota team europe were banned from the world rally championship for one year here on friday in a crushing ruling by the world council of the international automobile federation -lrb- fia - \\
		\bfseries{reference:} &  toyota are banned for a year \\
		\bfseries{direct:} & toyota banned from world rally championship \\
		\bfseries{nc:} & toyota europe banned from world rally championship for one year \\
		\midrule
		\bfseries{Example 4:} & \\
        \bfseries{source:} & oil prices roared higher towards \#\# dollars on monday as equity markets surged on government action aimed at tackling a severe economic downturn . \\
		\bfseries{reference:} & oil prices soar towards \#\# dollars\\
		\bfseries{direct:} & oil prices jump towards \#\# dollars \\
		\bfseries{nc:} & oil prices climb towards \#\# dollars \\
		\midrule
		\midrule
		& \bfseries{Translation} \\
		\bfseries{Example 5:} & \\
		\bfseries{source:} & \begin{CJK*}{UTF8}{gbsn}
		欧盟\ 和\ 美国\ 都\ 表示\ 可以\ 接受\ 这\ 一\ 妥协\ 方案\ 。
        \end{CJK*}  \\
        \bfseries{reference:} & both the eu and the us indicated that they can accept this plan for a compromise . \\
        \bfseries{direct:} & the eu and the united states indicated that it can accept this compromise .\\
        \bfseries{nc:} & the european union and the united states have said that they can accept such a compromise plan .\\
		\midrule
		\bfseries{Example 6:} & \\
		\bfseries{source:} & \begin{CJK*}{UTF8}{gbsn}
		那么\ 这些\ 这个\ 方面\ 呢\ 是\ 现在\ 警方\ 调查\ 重点\ 。
        \end{CJK*} \\
		\bfseries{reference:} & well , this is the current focus of \underline{police investigation} .  \\
		\bfseries{direct:} & these are present at the current \underline{police} . \\
		\bfseries{nc:} & then these are the key to the current \underline{police investigation} .  \\
		\midrule
		\bfseries{Example 7:} & \\
		\bfseries{source:} & \begin{CJK*}{UTF8}{gbsn}
		双方\ 有可能\ 就此\ 问题\ 在\ 下周\ 进行\ 磋商\ 。
        \end{CJK*} \\
		\bfseries{reference:} & the two sides may conduct negotiations on this issue next week .  \\
		\bfseries{direct:} & the two sides may hold consultations on next week . \\
		\bfseries{nc:} & the two sides are likely to hold consultations on this issue next week . \\
		\midrule
		\bfseries{Example 8:} & \\
		\bfseries{source:} & \begin{CJK*}{UTF8}{gbsn}
		那么\ 在\ 这个\ 问题\ 上\ ,\ 伊朗\ 现在\ 态度\ 比较\ 强硬\ ,\ 而\ 美国\ 的\ 态度\ 更为\ 强硬\ 。
        \end{CJK*} \\
		\bfseries{reference:} & well , iran 's attitude is now quite firm on this issue , while the us takes an even firmer attitude .  \\
		\bfseries{direct:} & on this issue , iran 's attitude is quite hard and the attitude of the united states is still tougher . \\
		\bfseries{nc:} & then , on this issue , iran has now taken a tougher attitude toward it . however , the attitude of the united states is even harder . \\
		\bottomrule
	\end{tabular}
	\caption {Example outputs on the test set from the direct model and noisy channel model for the summarisation task and machine translation.}
	\label{example}
\end{table}

\end{document}